\documentclass[sigconf]{acmart}
\AtBeginDocument{%
  }

\acmYear{2025}\copyrightyear{2025}
\setcopyright{acmlicensed}
\acmConference[MobiHoc '25]{International Symposium on Theory, Algorithmic Foundations, and Protocol Design for Mobile Networks and Mobile Computing}{October 27--30, 2025}{Houston, TX, USA}
\acmBooktitle{International Symposium on Theory, Algorithmic Foundations, and Protocol Design for Mobile Networks and Mobile Computing (MobiHoc '25), October 27--30, 2025, Houston, TX, USA}
\acmDOI{10.1145/3704413.3764420}
\acmISBN{979-8-4007-1353-8/25/10}

\usepackage{enumitem}
\usepackage{multirow}
\usepackage{xspace}

\newcommand{\systemname}{{mmExpert}\xspace}

\begin{document}

\title{mmExpert: Integrating Large Language Models for Comprehensive mmWave Data Synthesis and Understanding}

\author{Yifan Yan$^1$, Shuai Yang$^1$, Xiuzhen Guo$^1$ $^*$, Xiangguang Wang$^1$, Wei Chow$^1$, \\ Yuanchao Shu$^1$, Shibo He$^1$}
\thanks{$^*$Xiuzhen Guo is the corresponding author.}
\affiliation{%
  \institution{$^1$Zhejiang University}
  \city{Hangzhou}
  \state{Zhejiang}
  \country{China}
}

\renewcommand{\shortauthors}{Yan et al.}

\begin{abstract}
Millimeter-wave (mmWave) sensing technology holds significant value in human-centric applications, yet the high costs associated with data acquisition and annotation limit its widespread adoption in our daily lives. Concurrently, the rapid evolution of large language models (LLMs) has opened up opportunities for addressing complex human needs. This paper presents \systemname, an innovative mmWave understanding framework consisting of a data generation flywheel that leverages LLMs to automate the generation of synthetic mmWave radar datasets for specific application scenarios, thereby training models capable of zero-shot generalization in real-world environments. Extensive experiments demonstrate that the data synthesized by \systemname significantly enhances the performance of downstream models and facilitates the successful deployment of large language models for mmWave understanding. 
\end{abstract}




\maketitle

\section{Introduction}
In recent years, there have been significant advancements in the field of wireless sensing, especially millimeter-wave (mmWave)~\cite{zhang2025terahertz, sun2023bifrost, jiang2021sense, xue2021mmmesh, zhang2023survey}. The intrinsic non-invasive and privacy-preserving features make mmWave radars well-suited for a broad spectrum of human-centric applications, including healthcare, security, and smart homes~\cite{liu2021overview, liu2020real, zhang2023survey, guo2024exploring}. 
As the sensed mmWave signal contains information on human activities, state-of-the-art methods resort to deep neural models for inference. However, due to limited training datasets, the performance of these methods degrades drastically when encountering unseen test cases (e.g., users and environments). 

To address this issue, recent studies have proposed synthetic data generation methods to augment the training data for mmWave radar-based human sensing tasks~\cite{ahuja2021vid2doppler, zhang2022synthesized, xue2023towards, chen2023rf, zhou2024text2doppler}. These methods leverage the channel model, simulator, or deep learning model to generate training data without requiring data collection experiments involving humans in the loop. As the rapid advancements in large language models (LLMs) show remarkable potential for understanding multi-modal data~\cite{xu2024pointllm}, the latest works have been striving to generate mmWave radar data directly from textual descriptions or prompts~\cite{chen2023rf, chi2024rf, zhou2024text2doppler}. The LLM-based methods show impressive performance in mitigating data scarcity by leveraging the inherent web-scale world knowledge. 

Although the LLM has shown its great potential to mitigate data scarcity, the power of the LLM has not been fully unleashed in mmWave sensing. Compared with traditional mmWave sensing tasks like classification or reconstruction, the mmWave understanding task fundamentally differs by requiring open-vocabulary reasoning (Table~\ref{tab:demo}) instead of providing discriminative results. And this reasoning capability is uniquely achievable by LLMs. 

\begin{figure*}[ht]
\centering
\includegraphics[width=0.98\linewidth]{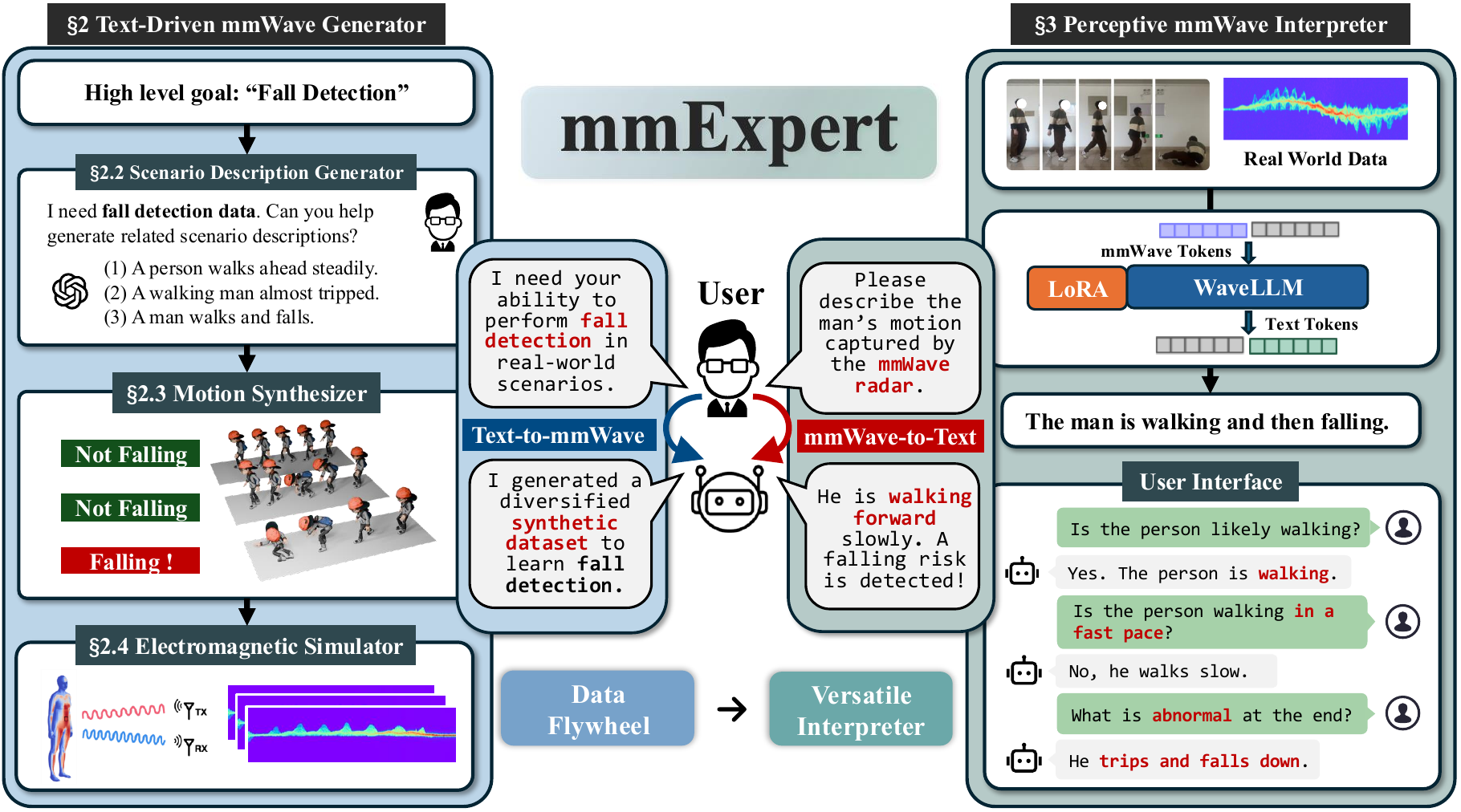}
\caption{Overview of the \systemname. \systemname comprises a data flywheel that generates mmWave-language data hierarchically and a versatile interpreter capable of understanding mmWave data across diverse contexts.}
\label{fig:system_overview}
\end{figure*}

In this paper, we present \textbf{\systemname}, a framework that enables text-to-mmWave cross-modal generation and mmWave-to-text sensing data understanding. As shown in Figure~\ref{fig:system_overview}, \systemname consists of two modules, i.e., ``text-to-mmWave'' and ``mmWave-to-text''. The ``text-to-mmWave'' module transforms textual descriptions into synthetic mmWave signals by leveraging LLM to automate the generation of synthetic mmWave radar datasets for required application scenarios.  Conversely, the ``mmWave-to-text'' module converts real-world mmWave signals into textual descriptions by utilizing open-vocabulary training and transferring pre-trained models from other domains. \systemname significantly reduces the reliance on extensive real-world data collection while equipping LLMs with the ability to interpret and understand mmWave sensing data, thereby enabling more robust and scalable mmWave sensing applications.

To achieve this, several non-trivial challenges need to be addressed. 

\noindent\textbf{(i) Decoupling real-world data dependency in the training stage.} 
Synthetic data often fails to capture the full characteristics of real-world scenarios, leading to potential discrepancies in model performance when deployed in practical applications. Furthermore, under these constraints, it is not feasible to train Neural Style Transfer (NST) models using any real-world data, and real-world data remains unavailable for the fine-tuning phase. To overcome these limitations, \systemname employs electromagnetic simulation to model the spatial propagation of mmWave signals, thereby accurately characterizing the physical properties of mmWave. Additionally, domain randomization techniques are utilized to design various random factors within the simulation system, which enhances the sim-to-real transfer without requiring real-world data. 

\noindent\textbf{(ii) Data curation pipeline for mmWave-text alignment.} Accurate and comprehensive textual annotation for mmWave data is an exceedingly expensive and labour-intensive process. Previous works have struggled to achieve high-quality mmWave-text alignment as their textual descriptions lack diversity and were restricted to a specific format~\cite{cao2024mmclip}. This limitation hinders the models' ability to integrate mmWave signals with free-form text. We address this challenge in \systemname by utilizing hierarchical action description prompts that encompass rich textual content to generate detailed action descriptions. The prompts allow \systemname to accommodate a wide range of actions without being restricted to a predefined set of categories, enhancing the flexibility and scalability of the dataset. Furthermore, \systemname employs a language-aligned pre-training phase, leveraging contrastive learning to establish a robust correspondence between the mmWave data and text descriptions. 

\noindent\textbf{(iii) Comprehensive benchmark of the mmWave interpreter.} Evaluating the mmWave Interpreter is particularly challenging due to its open-vocabulary text outputs, which traditional metrics cannot adequately assess. Moreover, the absence of a comprehensive test set further complicates the evaluation process. To address these challenges, we conduct extensive experiments using a diverse array of metrics, including traditional language metrics, machine learning-based metrics, and recent evaluations employing LLM. To better gauge the comprehensive understanding abilities of the interpreter, we also carefully curate a benchmark test set that specifically includes question-answering and captioning, which are pivotal for assessing core language skills. Specifically, this question-answering component utilizes natural language to pose questions that probe for fine-grained details about the observed actions and inquire into the possible intentions behind them.

Our contributions are summarized as follows:

\begin{itemize}[leftmargin=*,topsep=0.1em]
\item We propose a scalable and cost-effective framework for generating diverse mmWave signals using world knowledge from LLMs, eliminating the need for real-world data collection, which is often expensive and labor-intensive.

\item With the proposed training diagram and encoder design, we not only pioneer the exploration of scaling laws in mmWave signal modeling but also significantly improve performance, surpassing the strongest baseline by 19\% in classification tasks and reducing FID by 22\% in signal generation.

\item We design and train WaveLLM, a multi-modal large language model, exclusively on synthesized data. WaveLLM demonstrates strong zero-shot performance on real-world data, showcasing understanding and reasoning capabilities.
\end{itemize}

\section{Text-Driven mmWave Generator: Text-to-mmWave}

\subsection{Overview}
In this section, we introduce the design of the text-driven mmWave generator, the text-to-mmWave part of \systemname. It includes four modules. (1) \textbf{Scenario Description Generator} (\S\ref{subsec:scenario_cognition}) produces diverse and contextually accurate motion descriptions with user-defined textual prompts.
(2) \textbf{Motion Synthesizer} (\S\ref{subsec:motion_synthesis}) transform these motion descriptions into 3D human motion sequences by utilizing pre-trained motion generation models.
(3) \textbf{RF-Signal Electromagnetic Simulation} (\S\ref{subsec:electromagnetic_simulation}) applies physical electromagnetic principles to simulate radar signal interactions with the generated human models, yielding realistic radar data.
(4) \textbf{Sim-to-Real Domain Randomization} (\S\ref{subsec:domain_randomization}) further introduces variability in parameters such as radar angles, body segmentation, antenna pattern, background noise, and nonlinearity scaling to enhance the generalization and robustness of the system in real-world scenarios.

\subsection{Scenario Description Generator}
\label{subsec:scenario_cognition}

\systemname interprets the user's scenario description and generates a range of diverse motion prompts. \systemname utilizes world knowledge and the strong reasoning ability of the LLM. This enables \systemname to recognize not only explicit actions but also to infer implicit behaviors and possible outcomes based on typical situational dynamics. With prompt engineering and its own rich knowledge, the LLM generates accurate descriptions for precise text-to-motion translation and disassembles complex actions into manageable units.

To generate a high-quality text-mmWave paired synthetic dataset, two key considerations must be addressed: accurately defining the characters' actions and ensuring diversity in the descriptions. \systemname considers the following aspects: 

\noindent \textbf{Syntactic and synonym substitution.} \systemname generates motion description prompts using diverse syntactic structures, such as different tenses and inversion patterns, to emphasize aspects of the action while also drawing from a wide range of synonyms to enhance description variety.

\noindent \textbf{Temporal relation of actions.}
The temporal relationship between actions is crucial for coherent motion descriptions. \systemname considers the sequence and repetition of actions, ensuring that each action is presented in an appropriate temporal context.

\subsection{Motion Synthesizer}
\label{subsec:motion_synthesis}

We employ an off-the-shelf, text-driven human motion generation model \cite{guo2024momask} to perform as the motion synthesizer. This model is structured as a sequence-to-sequence architecture, which accepts descriptive text prompts as input and generates corresponding motion sequences. These motion sequences are typically represented as three-dimensional human skeletal keypoints. To transform these keypoint sequences into detailed and realistic human surface representations, we fit them to the SMPL human surface model~\cite{bogo2016keep}. This fitting process converts the skeletal data into a comprehensive 3D mesh that accurately depicts the human body's surface.

However, motion data generated in the temporal domain can exhibit potential jitter and frame-to-frame inconsistencies, which may compromise the overall quality and realism of the motion. To mitigate these issues, we apply Gaussian filtering to the generated human surface models along the time dimension. This filtering process smooths out abrupt changes and reduces noise, resulting in more fluid and coherent motion sequences that better capture natural human movements.

\subsection{RF Signal Electromagnetic Simulation}
\label{subsec:electromagnetic_simulation}


For the most commonly used Frequency Modulated Continuous Wave (FMCW) radar, the IF signal is essentially an estimation of the distance and reflection intensity of all reflection points, which can be expressed as
\begin{equation}
\small
\label{eq:IF_signal_multi_path}
x_{\mathrm{IF}}(t)=\sum_{i=1}^{N} a_i e^{j2 \pi \tau_i \left(f_c  + S t \right)},
\end{equation}
where $N$ is the number of reflection points, $f_c$ is the carrier frequency of the radar signal, $S$ is the frequency sweep rate of the radar signal, $a_i$ is the amplitude of the $i$-th reflection point, and $\tau_i$ is the time delay of the $i$-th reflection point. 

We observe that the amplitude of the IF signal plays a crucial role in ensuring the authenticity of the synthesized mmWave radar data. The amplitude $a_i$ of a single idealized path can be calculated:
\begin{equation}
\label{eq:amplitude}
a_i =
\frac{\lambda}{4\pi} \cdot 
\underbrace{\frac{1}{R^2}}_{\text{Path loss}} \cdot
\underbrace{\vphantom{\frac{1}{R^2}}\sqrt{\mathbf{C}_\mathrm{tx}\mathbf{C}_\mathrm{rx}}}_{\text{Antenna loss}} \cdot 
 \underbrace{\vphantom{\frac{1}{R^2}}\Gamma \sqrt{dA^\prime \mathbf{f}_{\mathrm{s}}}}_{\text{Scattering loss}}, 
\end{equation}
where $\mathbf{C}_\mathrm{tx}$ and $\mathbf{C}_\mathrm{rx}$ are the gains of the transmitting and receiving antennas after accounting for radar pattern, $\lambda$ is the wavelength of the millimeter wave signal, $R$ is the distance between the radar and the target, $d A^\prime$ is the projection area of the scattering surface, $\Gamma$ is the scattering coefficient, and $\mathbf{f}_{\mathrm{s}}$ is the scattering pattern function.  


The above Equation~\ref{eq:amplitude} inspires us to incorporate the variations in reflection intensity caused by path loss, antenna loss, and scattering loss into the signal simulation:

\noindent \textbf{(1) Antenna loss} represents inefficiencies in signal transmission and reception due to imperfections in the antenna design, leading to signal degradation.

\noindent \textbf{(2) Path loss} refers to the reduction in signal strength as electromagnetic waves propagate through space, which varies with the distance between the transmitter and the reflector.

\noindent \textbf{(3) Scattering loss} arises from the interaction of electromagnetic waves with the detection target, causing the signal to scatter in multiple directions, further reducing the reflected signal strength.

By incorporating these factors into the signal simulation, we can more accurately model the real-world behavior of radar signals and improve the overall realism and reliability of the simulation outcomes.

We adopt the mesh-tracking method to simulate the radar signal instead of sending uniform rays through space. Our approach only targets the visible facets of the human mesh model from the radar's perspective. 

To accurately model the micro-Doppler effect, we interpolate the mesh's position between keyframes to derive precise surface velocities. This technique allows us to perform computationally intensive ray-tracing at a lower frame rate (e.g., 20 fps) while still generating high-resolution, millisecond-level micro-Doppler data. 

\subsection{Sim-to-Real Domain Randomization}
\label{subsec:domain_randomization}

\begin{figure}[tp]
\centering
\includegraphics[width=0.95\linewidth]{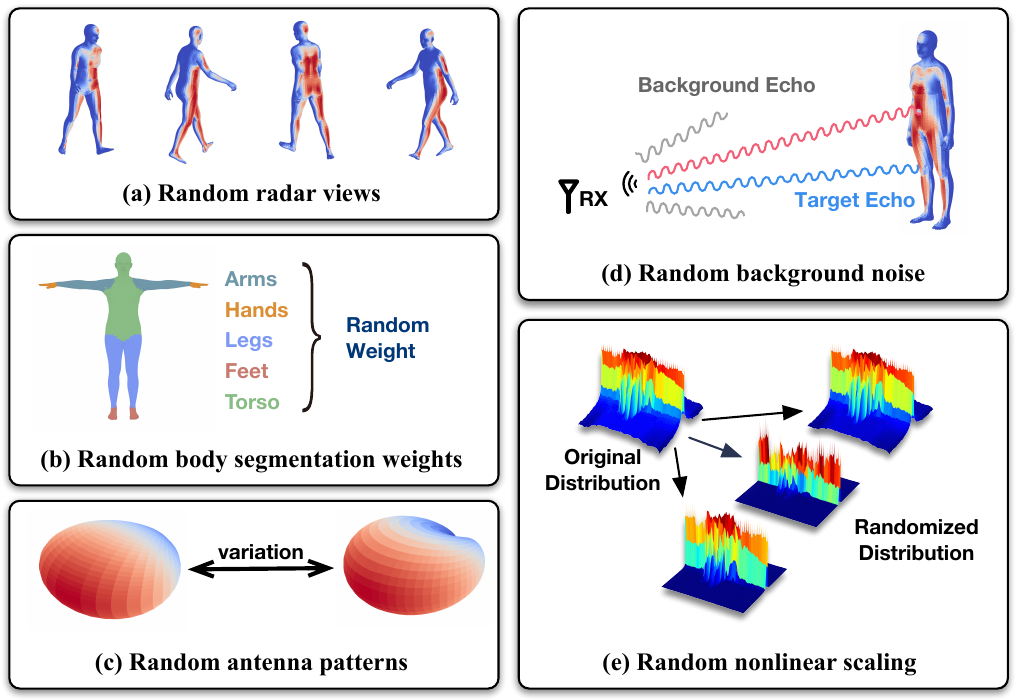}
\captionsetup{skip=1em}
\caption{Illustration of domain randomization factors. }
\label{fig:domain_adaption}
\end{figure}

To overcome the sim-to-real challenge, domain adaptation methods are essential to enable a model to generalize from synthetic data distributions to real-world data distributions. Previous works~\cite{chen2023rf, ahuja2021vid2doppler, deng2023midas++, zhang2022synthesized, zhou2024text2doppler} utilize the neutral style transfer method and the
fine-tuning method to improve the sim-to-real transfer performance. However, both methods rely on real-world data, which is costly and time-consuming to collect. In contrast, \systemname operates under a zero-shot generalization condition, which means that our system does not depend on real-world data during the training stage. 

The domain randomization technique~\cite{tobin2017domain} introduces sufficient random variation into the simulated training data, allowing the model to learn robust features that are invariant to distribution shifts between the synthetic data and the real-world data. In our implementation, we apply the following domain randomization variations as shown in Figure~\ref{fig:domain_adaption}.

\noindent \textbf{(a) Random radar views.}
We randomly alter the radar view angle to capture radar signals from different perspectives. This variation enables the downstream model to learn invariant features of human motion, regardless of the observation angle.

\noindent \textbf{(b) Random body segmentation weights.}
The human mesh is divided into segments corresponding to body parts based on anatomical structure. Random weights are assigned to each part to simulate varying reflection intensities, emulating the diverse material properties of the human body.

\noindent\textbf{(c) Random antenna patterns.}
Due to the complexity of accurately modeling radar antenna characteristics, we randomly adjust the beam width of the simulated radar in both azimuth and elevation within a predefined range. This is achieved by assigning parameters to the antenna pattern function as described in \S\ref{subsec:electromagnetic_simulation}.

\noindent \textbf{(d) Random background noise.}
Gaussian white noise is added to the synthesized mmWave radar data to simulate thermal noise inherent in radar systems. Additionally, to emulate static background noise encountered in real-world scenarios, we generate background echoes with random energy intensities during simulation.

\noindent \textbf{(e) Random nonlinearity scaling.}
In typical data preprocessing, radar signals undergo dB conversion to compress the dynamic range. However, the logarithm operation is extremely sensitive to the input data distribution, which introduces nonlinearity. We apply stochastic nonlinearity scaling to enhance the diversity of the distribution of the synthesized training data.

\section{Perceptive mmWave Interpreter: mmWave-to-Text}
\label{sec:interpreter}

\subsection{Overview}

In this section, we introduce the Perceptive mmWave Interpreter, the mmWave-to-Text component of \systemname. The interpreter is designed to convert mmWave radar signals into meaningful textual descriptions, thereby enhancing downstream perception tasks. The architecture consists of four main modules. (1) \textbf{mmWave Signal Feature Extractor} (\S\ref{subsec:mmwave_encoder}) utilizes a transformer-based encoder to capture characteristics of mmWave signals, providing a comprehensive set of features essential for semantic interpretation. (2) \textbf{Sentence Feature Extractor} (\S\ref{subsec:sentence_encoder}) employs a BERT-based model to generate rich semantic embeddings from textual inputs, ensuring effective representation of natural language descriptions. (3) \textbf{Language-Aligned Pre-training}  (\S\ref{subsec:mmwave_text_alignment_pretraining}) adopts a contrastive learning approach inspired by CLIP to align mmWave signal features with corresponding text embeddings, creating a cohesive semantic framework that bridges the gap between radar data and language. (4) \textbf{mmWave Interpreter: WaveLLM} (\S\ref{subsec:wavellm}) integrates the aligned features into a large language model (LLM) framework, enabling the generation of contextually accurate and semantically rich textual descriptions based on mmWave signal inputs.

\subsection{mmWave Signal Feature Extractor}
\label{subsec:mmwave_encoder}

\begin{figure}[tp]
\centering
\includegraphics[width=\linewidth]{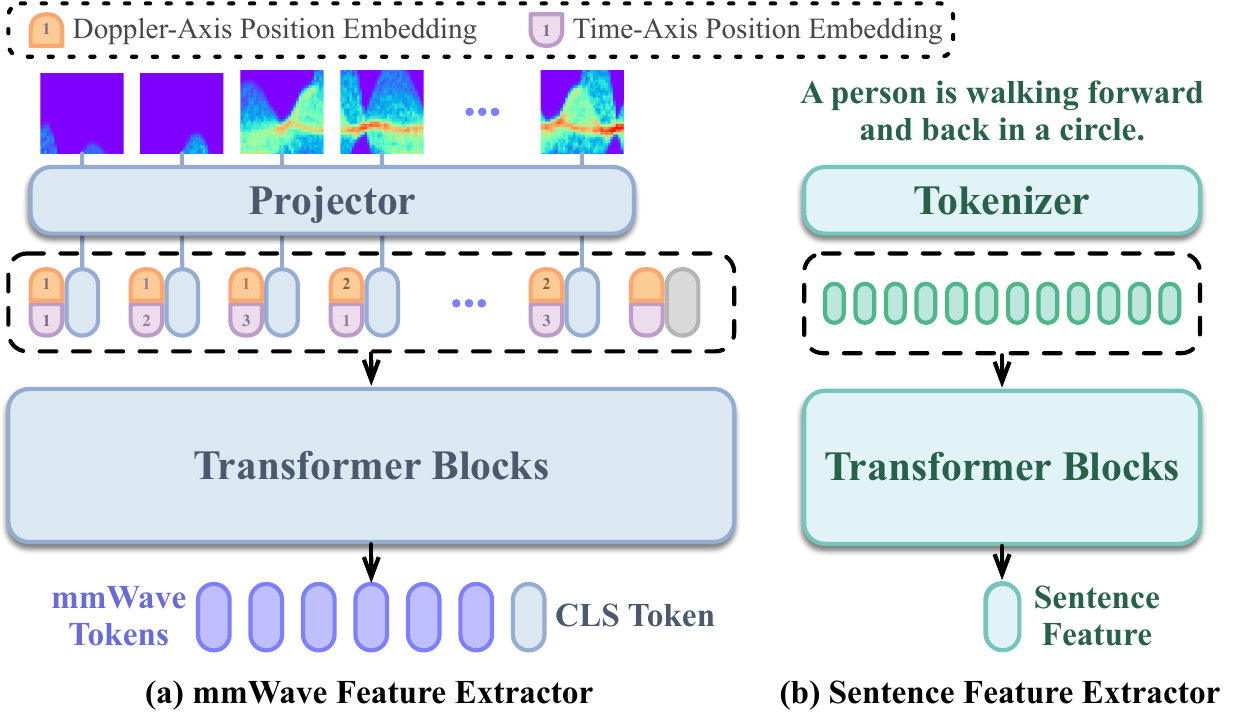}
\captionsetup{skip=0em}
\caption{Architecture of the encoders. The mmWave encoder extracts motion features from the signal, and the sentence encoder extracts semantic features from the text.}
\label{fig:encoder}
\end{figure}

We leverage a transformer-based encoder to capture temporal information from the micro-Doppler spectrum, which provides a compact and informative representation of human motion. This module processes the input spectrogram data as a 2D matrix $\boldsymbol{I} \in \mathbb{R}^{H \times W}$, where $H$ represents the temporal axis and $W$ corresponds to the Doppler frequency axis, representing the target’s radial velocity. The encoder architecture is shown in Figure~\ref{fig:encoder} (a). 

To extract meaningful features, the input spectrogram is partitioned into a sequence of flattened 2D patches $x \in \mathbb{R}^{N \times \left(P^2\right)}$, with $P$ denoting the patch size and $N$ the number of patches. These patches are then projected into a latent feature space of dimension $D$. To effectively preserve the contextual information along both the temporal and Doppler axes, we incorporate learnable positional embeddings for each dimension separately. 

Next, we utilize a multi-layer transformer encoder to extract effective global features from the input. By incorporating a learnable class token, the encoder aggregates comprehensive information, enabling the module to summarize the overall spectrogram information. 

The output of this module includes both a global feature embedding derived from the class token and a sequence of token embeddings $z \in \mathbb{R}^{(N+1) \times D}$. The global feature is aligned with text embeddings during the contrastive learning process (\S\ref{subsec:mmwave_text_alignment_pretraining}), while the sequence embeddings retain temporal motion details for subsequent semantic modeling tasks. Together, these features comprehensively represent the temporal characteristics essential for mmWave semantic interpretation. 

\subsection{Sentence Feature Extractor}
\label{subsec:sentence_encoder}

The Sentence Feature Extractor is a BERT model~\cite{reimers2019sentence} meticulously designed to generate rich semantic representations from textual inputs, specifically at the sentence level, as shown in Figure~\ref{fig:encoder} (b). Built upon a transformer-based architecture, it effectively captures intricate contextual dependencies within text through its core, which consists of multiple transformer layers utilizing self-attention mechanisms.

The text extractor begins by embedding input tokens into a high-dimensional feature space, capturing essential syntactic and semantic information. Positional encodings are added to preserve the sequential structure of the text. The processed tokens are then passed through a transformer encoder, similar to the previously described mmWave encoder (\S\ref{subsec:mmwave_encoder}), which aggregates global sentence-level features. 

At the final stage, a pooling mechanism is applied to generate a compact sentence-level embedding. Specifically, the model aggregates information from all token representations, effectively summarizing the overall semantic content of the input. This representation serves as the global feature for subsequent alignment with the mmWave-based motion features during the contrastive learning phase (\S\ref{subsec:mmwave_text_alignment_pretraining}). 

To enhance the capability, the Sentence Feature Extractor is pre-trained on an extensive corpus of diverse natural language tasks. Such robust pre-training enables it to capture the full semantic essence of sentences accurately. Consequently, it can extract rich, context-aware representations that generalize proficiently across various language-understanding tasks.

\subsection{Language-Aligned Pre-training}
\label{subsec:mmwave_text_alignment_pretraining}

We utilize a contrastive learning approach inspired by the CLIP~\cite{radford2021learning} method to align mmWave signal features with text features. This alignment is essential for integrating mmWave motion features with semantic textual content, ultimately enriching the understanding of scene and event descriptions.

The contrastive learning framework pairs mmWave signal features, extracted via the mmWave Signal Feature Extractor, with sentence embeddings generated by the Sentence Feature Extractor. During training, the model learns to maximize the similarity of corresponding pairs (i.e., the mmWave feature corresponding to a specific textual description) while minimizing the similarity between non-corresponding pairs. This dual-objective approach encourages the model to discern fine-grained relationships between motion and text, bridging the gap between these modalities.

Specifically, we use cosine similarity to evaluate the distance between multi-modal features in the latent space, enabling seamless comparisons between mmWave and textual representations. The model is then trained using the InfoNCE loss~\cite{oord2018representation}, defined as:
\begin{equation}
\footnotesize
\mathcal{L}_{\text{CLIP}} = -\frac{1}{N} \sum_{i=1}^{N} \log \left( \frac{\exp(\boldsymbol{v}_i \cdot \boldsymbol{t}_i / \tau)}{\sum_{j=1}^{N} \exp(\boldsymbol{v}_i \cdot \boldsymbol{t}_j / \tau) + \sum_{j=1}^{N} \exp(\boldsymbol{t}_i \cdot \boldsymbol{v}_j / \tau)} \right), 
\end{equation}
where $\boldsymbol{v}_i$ and $\boldsymbol{t}_i$ represent the mmWave feature and its corresponding text feature embedding for the $i$-th sample, and $\tau$ is a temperature parameter that scales the logit outputs. This loss function minimizes the distance between positive pairs and maximizes the distance between negative pairs, fostering a robust semantic alignment. By aligning the mmWave features with text features, we enable the model to map the disparate data types into a coherent semantic space.

\begin{figure*}[tp]
\centering
\includegraphics[width=\linewidth]{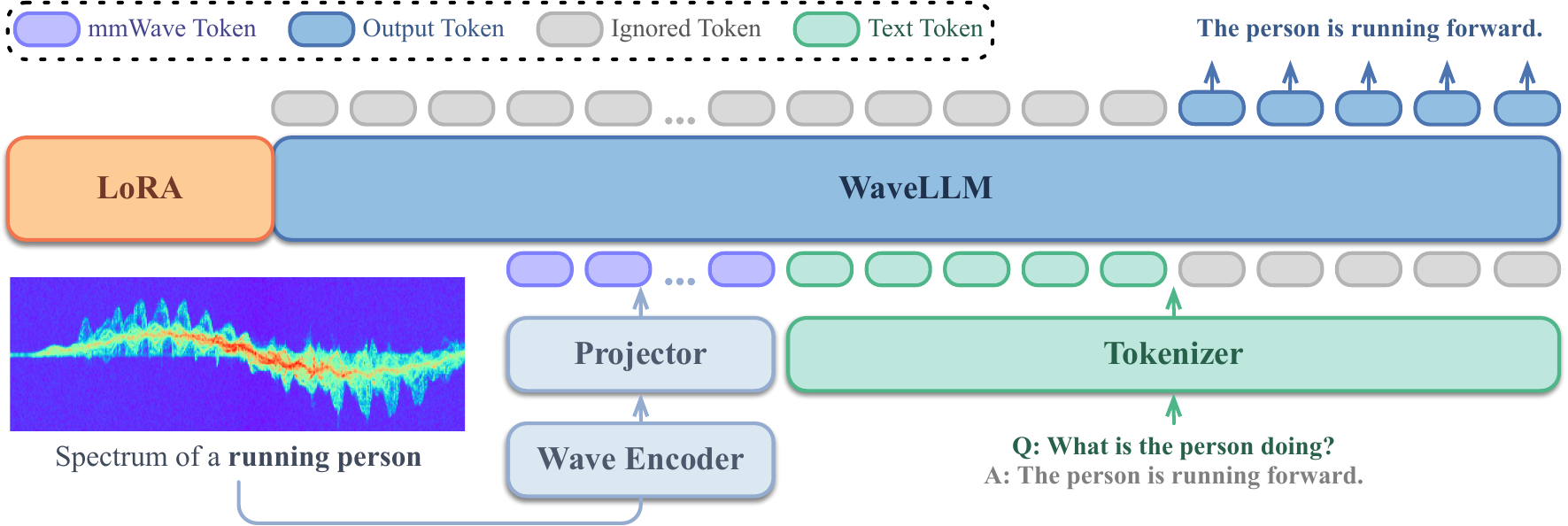}
\captionsetup{skip=0.5em}
\caption{Architecture of WaveLLM. The wave encoder extracts features from the signal, and the projector maps these features to the LLM's embedding space. Leveraging multi-modal context, the LLM generates the answer.}
\label{fig:wavellm}
\end{figure*}

\subsection{mmWave Interpreter: WaveLLM}
\label{subsec:wavellm}
To further unlock the power of mmWave semantic understanding, we present WaveLLM, a generative model designed to understand complex multi-modal contexts from mmWave signals and language. The model comprises three main components: a pre-trained mmWave feature extractor $f_{E}$, detailed in \S~\ref{subsec:mmwave_encoder}, a multi-modal projection layer $f_{P}$, and a pre-trained large language model (LLM) backbone $f_{LLM}$.

To enable WaveLLM to handle both mmWave signal features and language features effectively, we enhance the pre-trained LLM's capabilities. This is accomplished by strategically combining the two types of data. Specifically, we take the features extracted from mmWave signals and align them with corresponding language features. Before these features are combined, a shallow MLP (multi-layer perceptron) based projection layer is used, denoted as $f_{P}$. This layer plays a crucial role in converting mmWave features into a form that can be easily integrated with language features. Once the conversion is complete, the model concatenates these aligned mmWave and language features. This combined input is then fed into the LLM, which processes it to understand the multi-modal content. The role of the LLM is to sequentially predict the next token in the output sequence in an auto-regressive manner. As the user inputs a sequence of mmWave signals along with a corresponding question, the WaveLLM responds based on the underlying human motion depicted in the signals. 

To preserve the generality of the pre-trained LLM while enabling it to comprehend and reason with mmWave signals, we apply Low-Rank Adaptation (LoRA)~\cite{hu2022lora}. 

LoRA modifies the forward pass of linear layers in the Transformer without fully fine-tuning them. Instead, it introduces trainable low-rank matrices, altering the forward pass as follows:
\begin{equation}
h = W_0 x + \Delta W x = W_0 x + BA x,
\end{equation}
where $h \in \mathbb{R}^d$ and $x \in \mathbb{R}^k$ are the output and input of the layer, respectively; here, $W_0$ denotes the original weight matrix, which remains frozen during training. The trainable weights are $A \in \mathbb{R}^{r \times k}$ and $B \in \mathbb{R}^{d \times r}$, where $r \ll \min(d, k)$. 

By utilizing the efficient design of WaveLLM, we effectively train our multi-modal LLM to offer robust understanding capabilities, adeptly interpreting and reasoning across the mmWave and text data spaces.

\section{Implementation Details}
\label{sec:implementation}

\subsection{Radar Configuration} 
\begin{figure}[bp]
\centering
\includegraphics[width=0.98\linewidth]{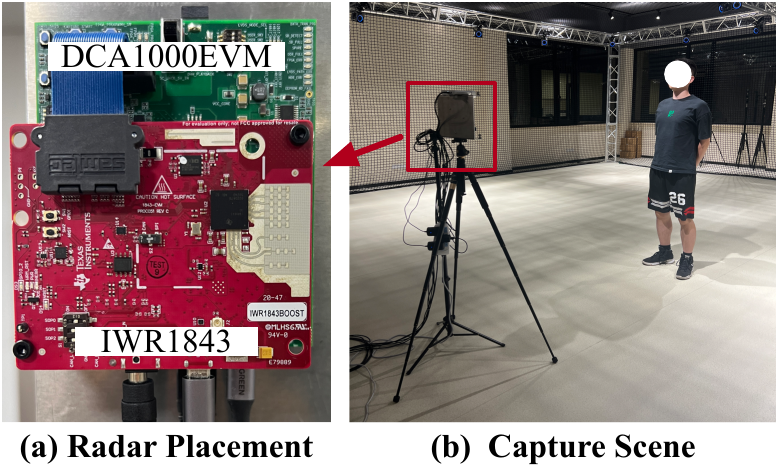}
\captionsetup{skip=0em}
\caption{Data capture system. The collected data is utilized exclusively during the testing phase and is not employed at any stage of training.}
\label{fig:radar_setup}
\end{figure}

The mmWave radar parameters in simulation and real-world data capture systems are synchronized. For the real-world hardware system, we utilize an IWR1843BOOST radar sensor equipped with three transmit antennas, four receive antennas, and a DCA1000EVM evaluation board by Texas Instruments for real-time data capture and streaming. The radar operates at a frequency of 77 GHz and bandwidth of 4 GHz, providing a range resolution of about 4.3 cm. The frame rate of the radar is set to 50fps, and each radar frame consists of 128 chirps. The radar is placed at a height of 1 meter, and the distance between the radar and the human subject ranges from 1 to 5 meters. To fully capture the whole motion of the human body, we put the radar sideways, which reduces the vertical energy attenuation, as illustrated in Figure~\ref{fig:radar_setup}. 

\begin{table*}[htbp]
\footnotesize
\centering
\captionsetup{skip=0em}
\caption{Real-world deployment demo of WaveLLM. All responses from WaveLLM are derived directly from the multi-modal LLM's inference on mmWave signal.}
\label{tab:demo}
\begin{tabular}{@{}
  p{0.1\textwidth} @{\hspace*{0.02\textwidth}}
  p{0.28\textwidth} @{\hspace*{0.02\textwidth}}
  p{0.28\textwidth} @{\hspace*{0.02\textwidth}} 
  p{0.28\textwidth}@{}}
\toprule
\textbf{Scenario} & 
\begin{minipage}{\linewidth}
\centering
\includegraphics[width=\linewidth]{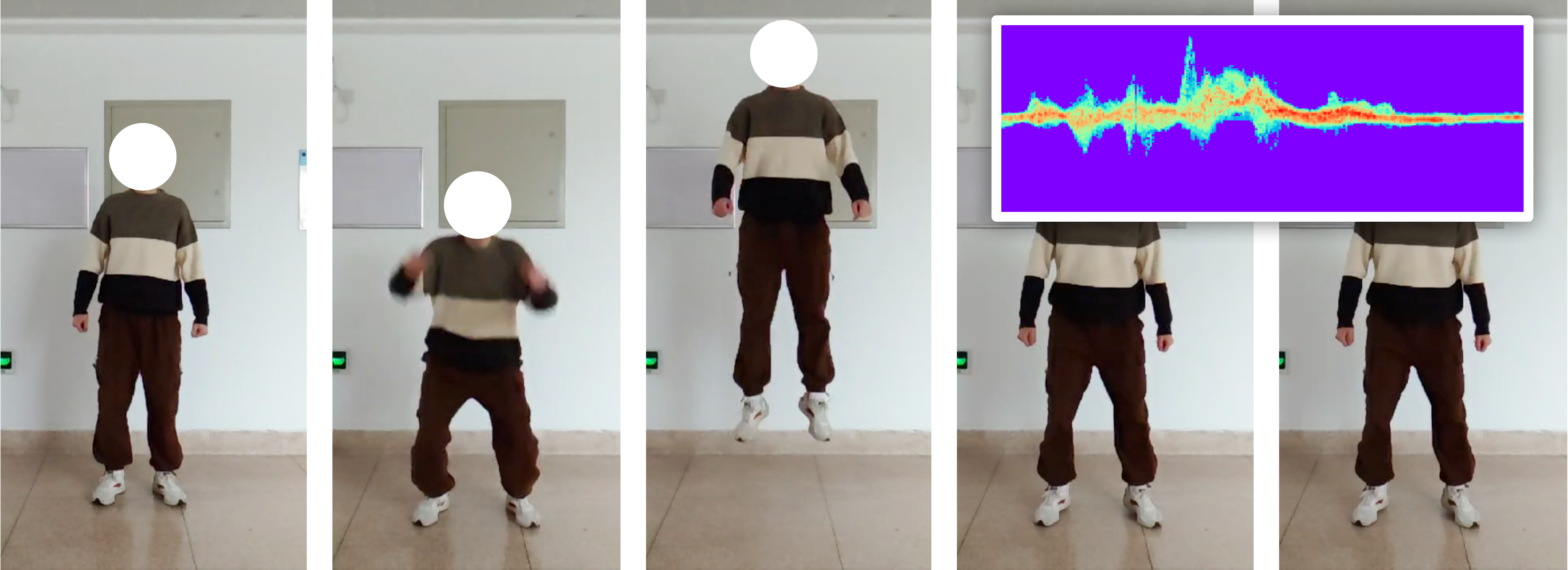}
\end{minipage} & 
\begin{minipage}{\linewidth}
\centering
\includegraphics[width=\linewidth]{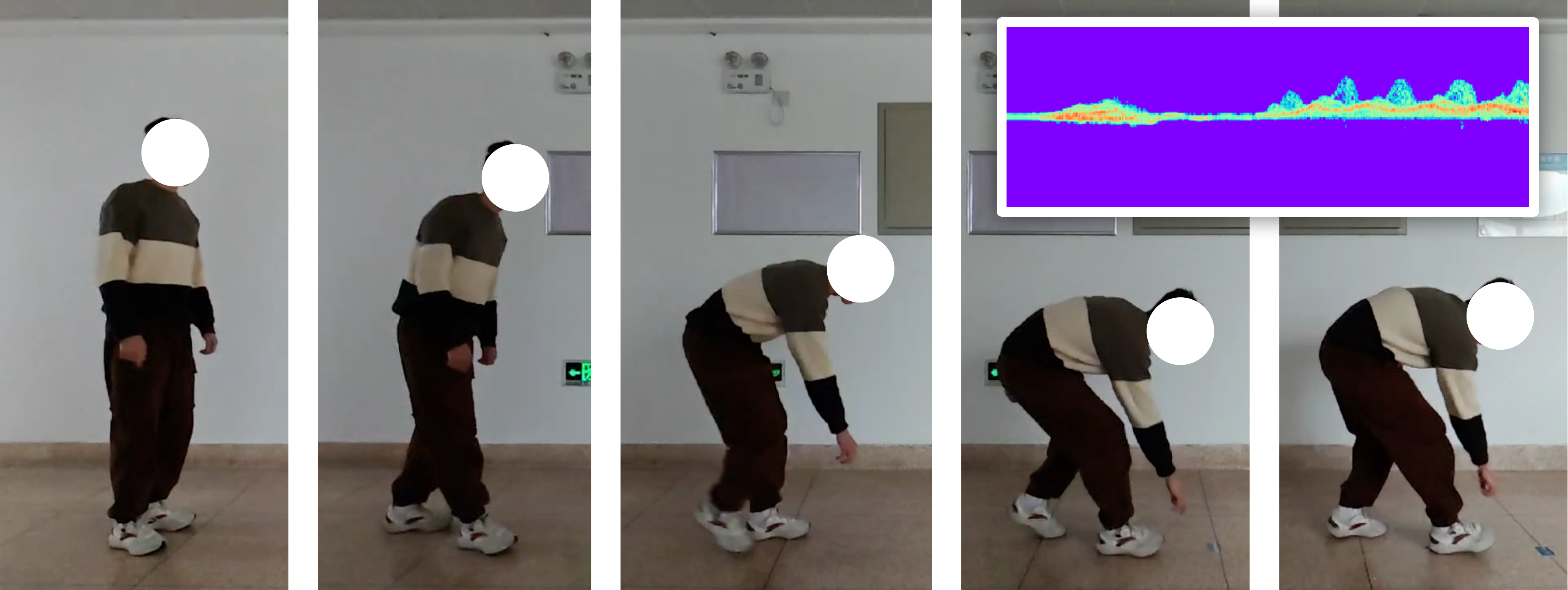}
\end{minipage} & 
\begin{minipage}{\linewidth}
\centering
\includegraphics[width=\linewidth]{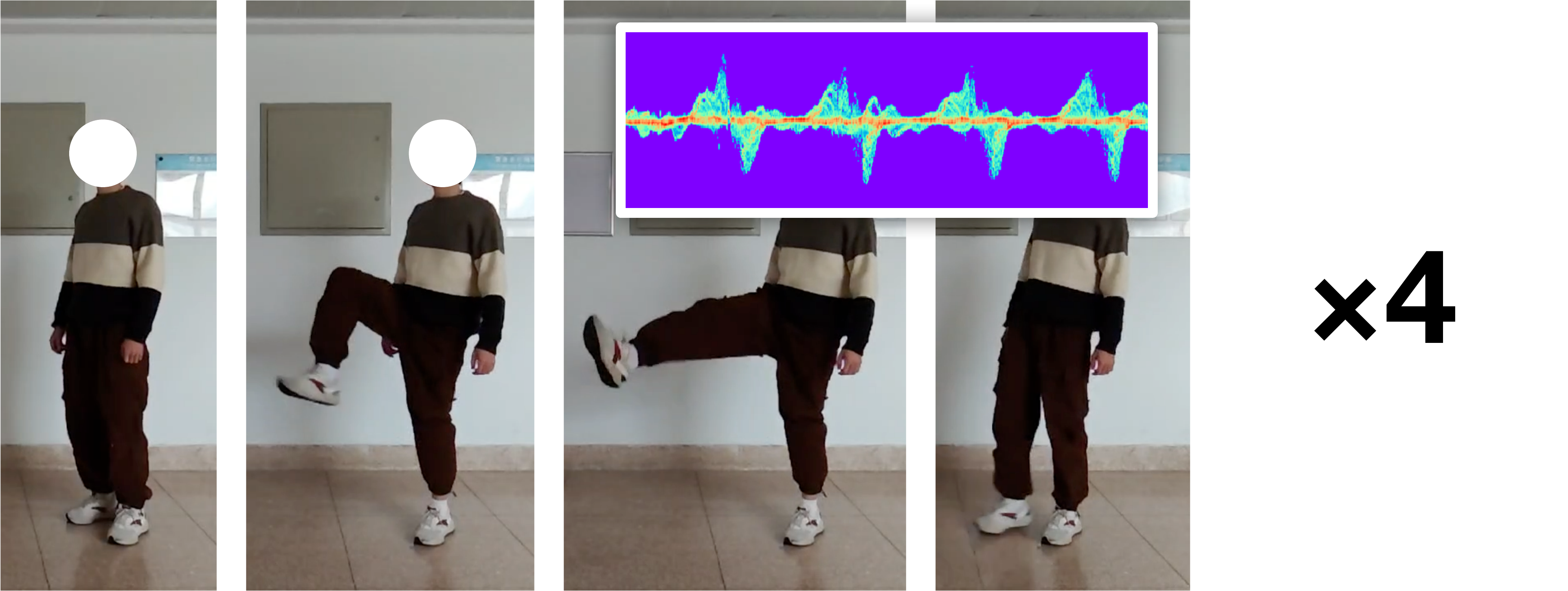}
\end{minipage} \\

\midrule

\medskip\medskip\medskip\medskip\medskip
\medskip\medskip\medskip\medskip\medskip
\medskip\medskip\medskip\medskip\medskip
\textbf{Q\&A} & 

\textbf{Predicted Caption:} The person is jumping up and down in place.

\medskip\medskip

\textbf{User:} Does the person stand still?

\textbf{WaveLLM:} No, the person is actively moving and jumping. 

\medskip\medskip

\textbf{User:} Is the person jumping in rapid succession or jumping hard and high?

\textbf{WaveLLM:} The person is jumping hard and high.

\medskip\medskip

\textbf{User:} Does the person stop jumping at the end?

\textbf{WaveLLM:} Yes, the person stops jumping at the end.
& 
\textbf{Predicted Caption:} The person bends over, then walks forward. 

\medskip\medskip

\textbf{User:} How many actions are involved according to the wave signal?

\textbf{WaveLLM:} Two actions, bending down and walking.

\medskip\medskip

\textbf{User:} What might the person want to do? 

\textbf{WaveLLM:} The person might want to pick something up from the ground.

\medskip\medskip

\textbf{User:} Does the man walk fast?

\textbf{WaveLLM:} No, he walks slowly.
& 
\textbf{Predicted Caption:} The person kicks in place multiple times. 

\medskip\medskip

\textbf{User:} Is the person likely to do sports?

\textbf{WaveLLM:} Yes, the person is likely engaging in a sport that involves kicking. 

\medskip\medskip

\textbf{User:} What kind of sport might he be playing?

\textbf{WaveLLM:} He could be playing soccer or martial arts. 

\medskip\medskip

\textbf{User:} How many kicks does this person perform?

\textbf{WaveLLM:} The person performs four kicks.

\\
\bottomrule
\end{tabular}
\end{table*}

\subsection{Data Curation} 

The training data consists of the HumanML3D~\cite{guo2022generating} dataset and the synthesized mmWave signals generated by the Scenario Description Generator. The HumanML3D dataset is a publicly available 3D human motion dataset that provides diverse motion capture data across various human activities. Meanwhile, HumanML3D provides high-quality text annotations for each motion sequence, which contain detailed descriptions of the motion characteristics. However, it is important to note that the categories included in HumanML3D are limited and do not encompass all complex actions encountered in daily life. Therefore, there remains a need to employ text-to-mmWave synthesis systems to augment this data effectively. 

We utilize the Scenario Description Generator introduced in \S\ref{subsec:scenario_cognition} to generate description prompts for extended training data generation. These generated prompts are used to guide the motion generation and the mmWave signal synthesis process. Notably, we do not explicitly provide motion category labels during the prompt generation stage for scenario guidance. Instead, the downstream model is required to learn the correlations between open-vocabulary prompts and synthesized mmWave signals; subsequently, it selects the most relevant labels based on incoming mmWave signals during the inference stage. 

The real-world evaluation dataset consists of over 1000 real-world mmWave micro-Doppler spectra corresponding to human motion with high-quality human-verified text annotations, including class labels, detailed captions, and question-answerings. The benchmark is able to evaluate the encoder's open vocabulary classification ability and the LLM's multi-modal understanding ability. 

Each spectrum sample is a time-Doppler signal with a duration ranging from 6 to 12 seconds, captured at 50 fps. Our dataset is collected across two scenarios and includes 10 volunteers. It comprises 15 action categories in total: 9 basic actions (e.g., walking, running, bending over) and 6 complex actions, which are combinations of the basic movements. Each action sequence varies in speed and frequency and corresponds to a unique text description.

\subsection{Model Training Detail}

We use a pre-trained Vision Transformer~\cite{dosovitskiy2020image} as the spectrum feature extractor, characterized by a patch size of 16. The model comprises 12 transformer layers, each with 12 attention heads and a hidden size of 768. For the sentence feature extractor, we employ Sentence-BERT~\cite{reimers2019sentence}, a pre-trained model specifically designed to generate semantically meaningful sentence embeddings. The model is based on the BERT architecture and consists of 6 hidden layers, each containing 12 attention heads and a hidden size of 384. 

During the CLIP training phase, we set the learning rate to $1 \times 10^{-4}$, and the temperature parameter is configured to 0.1. The model is trained for 50 epochs to ensure convergence and optimal performance. 

For the mmWave interpreter, we employ a compact yet robust language model backbone, Phi-3~\cite{phi3}, which consists of 3.8 billion parameters. In our primary analysis, we utilize LoRA~\cite{hu2022lora} with a rank of 8, where the trainable weight percentage is 0.43\%. 

\section{Evaluation Results} 
\label{sec:evaluation}

\subsection{Baselines}

\noindent\textbf{Baselines for data synthesis. } To illustrate the quality of the generated data, we choose the state-of-the-art publicly available text-based mmWave signal synthesis systems as the baseline systems for comparison. 

\begin{itemize}[leftmargin=*,topsep=0.1em, partopsep=0.1em, itemsep=0.1em]
\item \textbf{RFGen \cite{chen2023rf}}  utilizes a diffusion-based text-to-motion model to generate human mesh models. It proposes a physics-based ray tracing simulator for mmWave signal synthesis. RFGen further adopts another diffusion-based neural style transfer model to overcome the sim-to-real gap. 

\item \textbf{Text2Doppler \cite{zhou2024text2doppler}}  utilizes a transformer-based generative model to generate motion from text prompts \cite{guo2024momask}. It approximates the human body segments as ellipsoids. We use the signal synthesis pipeline in Text2Doppler for comparison. 
\end{itemize}

The classification models are trained using the same configuration, with the only difference being the training data generated by different synthesis systems. While RFGen offers an end-to-end generation pipeline, we directly input the motion description prompts and obtain the synthesized mmWave signals. For Text2Doppler, we use the same human motion data generated by our motion synthesizer module, since only the signal simulator is released. 

\noindent\textbf{Baselines for mmWave understanding. } To the best of our knowledge, \systemname is the first framework to integrate large language models (LLMs) for mmWave understanding. For comparison, we use WaveLLM against two types of baselines: non-semantic alignment training methods (Classification) and fundamental models (ResNet). 

\begin{figure*}[tp]
\centering
\includegraphics[width=\linewidth]{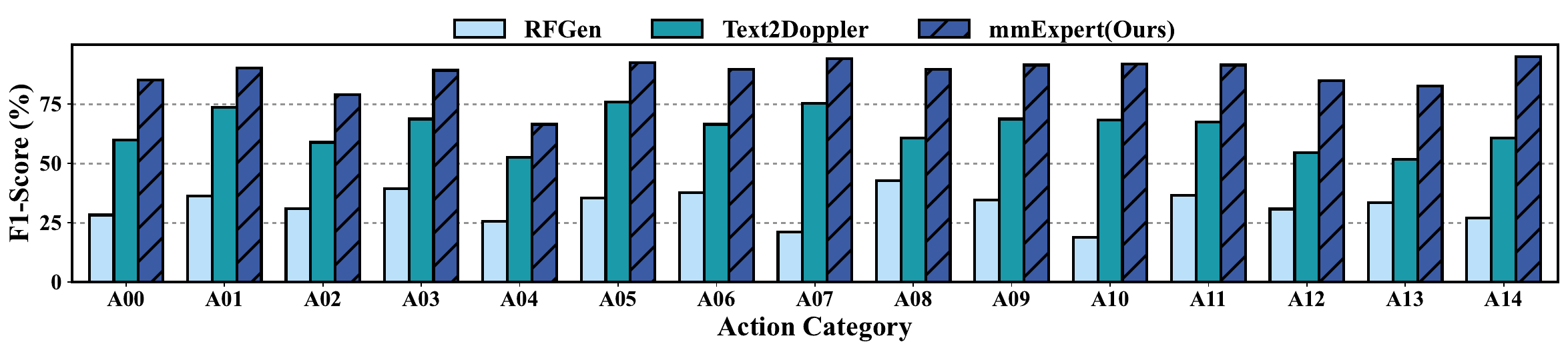}
\captionsetup{skip=0em}
\caption{Comparison of F1-scores for each action category with baselines. A00 to A14 refer to different action categories. The results indicate that the RFGen shows poor sim-to-real generalization ability, and our method surpasses Text2Doppler with a consistent performance gain.}
\label{fig:classification_results}
\end{figure*}

\begin{table*}[ht]
\small
\centering
\captionsetup{skip=0em}
\caption{WaveLLM is evaluated using GPT-based assessments and widely used language metrics. ``CLS'' represents classification as the training objective, while ``CLIP'' refers to contrastive learning. ``ResNet'' and ``ViT'' denote encoder architectures.}
\label{tab:llm_results}
\begin{tabular*}{\textwidth}{l@{\extracolsep{\fill}}cccccccc}
\toprule
\multirow{2}{*}[-0.5ex]{\textbf{Model}} & \multicolumn{3}{c}{\textbf{GPT Evaluation}} & \multicolumn{5}{c}{\textbf{Text-based Evaluation}} \\
\cmidrule(lr){2-4} \cmidrule(lr){5-9}
& \textbf{Correctness} & \textbf{Hallucination$\downarrow$} & \textbf{Precision} & \textbf{SBERT} & \textbf{SimCSE} & \textbf{BLEU-1} & \textbf{ROUGE-L} & \textbf{METEOR} \\
\midrule
WaveLLM-CLS-ResNet & 2073 & 889 & 70.0 & 69.0 & 67.5 & 41.5 & 45.3 & 21.0 \\
WaveLLM-CLS-ViT & 2049 & 913 & 69.2 & 69.4 & 67.7 & 42.3 & 45.7 & 21.3 \\
WaveLLM-CLIP-ResNet & 2105 & 857 & 71.1 & 69.8 & 68.6 & 41.9 & 46.0 & 21.3 \\
WaveLLM-CLIP-ViT & \textbf{2196} & \textbf{766} & \textbf{74.2} & \textbf{72.2} & \textbf{71.2} & \textbf{46.9} & \textbf{49.9} & \textbf{23.5} \\
\bottomrule
\end{tabular*}
\end{table*}

\subsection{Overall Performance}

\noindent\textbf{Zero-shot classification performance.} The results of the zero-shot generalization classification are presented in Figure~\ref{fig:classification_results}. The classification model (ViT encoder + MLP decoder), trained using the synthesized mmWave signals from our proposed \systemname, demonstrates outstanding performance across 15 human motion categories, achieving an impressive average accuracy of 84.6\%. To measure the model's ability to identify positive instances while minimizing false negatives, we calculate the F1-score for each category. 

Furthermore, we find that certain similar human actions exhibit slightly lower accuracies; nonetheless, \systemname consistently outperforms baseline methods in these cases. This underscores the effectiveness of \systemname in producing high-quality mmWave signals for real-world human activity recognition. 

We observed that RFGen's ray-tracing strategy based on depth images may cause abrupt changes in pixel-level depth during fast motion, thereby leading to periodic extension in the Doppler dimension and disrupting the time-Doppler signal. Meanwhile, Text2Doppler uses a simplified human model without sim-to-real design. Building on these observations, \systemname is designed to synthesize high temporal-resolution data by tracking the model's surface and employs a dedicated sim-to-real module to bridge the gap to real-world performance.

\noindent\textbf{WaveLLM understanding performance.} In this section, we primarily present the optimal configuration of WaveLLM. The test set assesses captioning and question-answering tasks, emphasizing the LLM's comprehension capabilities. We selected details from each description for point-based scoring. As shown in Table~\ref{tab:llm_results}, ViT consistently outperforms ResNet, and the CLIP training paradigm surpasses classification-based training. WaveLLM with ViT pre-trained using CLIP demonstrates improved accuracy in answer recognition and reduced hallucination, as indicated by GPT evaluation results, while also achieving superior overall language performance across various metrics. 

\subsection{Ablation Studies}

\noindent\textbf{Data scaling law.} The data scaling law~\cite{kaplan2020scaling,lin2024data} plays a pivotal role in understanding and optimizing the relationship between model performance and the amount of training data. To investigate this, we trained a CLIP classification model using both the human-labeled dataset (i.e., HumanML3D) and the synthesized dataset generated by \systemname from text descriptions. 

\begin{figure}[htbp!]
\centering
\includegraphics[width=\linewidth]{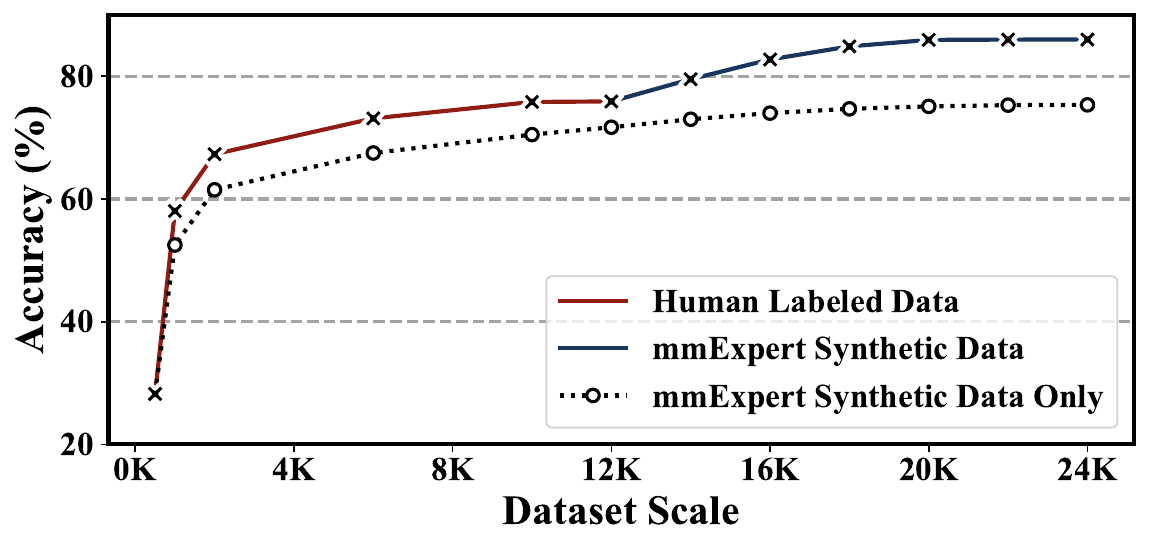}
\captionsetup{skip=0em}
\caption{Ablation of the extension dataset scale. When human-labeled data reaches saturation (red), data from \systemname (blue) further extends the boundary. }
\label{fig:scale}
\end{figure}

As shown in Figure~\ref{fig:scale}, the red curve represents the data scaling law derived from the human-labeled dataset, highlighting the critical trend that increased data generally leads to better performance. Notably, the blue curve showcases the scaling trend extended with our dataset, underscoring its ability to sustain consistent performance gains and further validate the significance of the scaling law in driving progress. This setup allows us to compare the impact of human-curated data versus synthetic data on model performance and to further demonstrate the robustness of the data scaling law. 

\begin{figure}[tp]
\centering
\includegraphics[width=\linewidth]{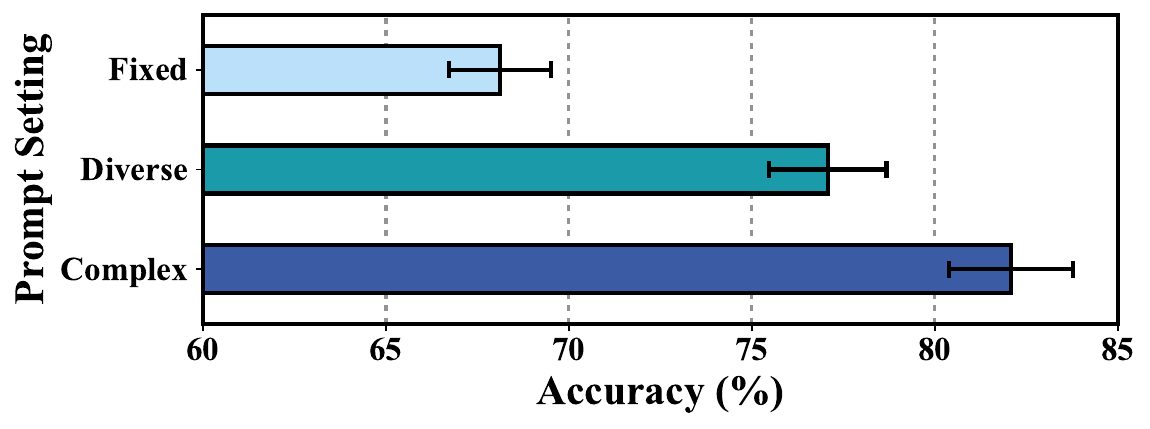}
\captionsetup{skip=0em}
\caption{Ablation of the description types. Legend: Fixed -- template-based descriptions; Diverse -- free-form descriptions; Complex --temporal action combination.}
\label{fig:prompts}
\end{figure}

\noindent\textbf{Impact of scenario-guiding prompts on dataset quality.} We evaluate the effect of scenario-guiding prompts on the quality of generated datasets through an ablation study, wherein we modify the prompts used to synthesize mmWave signals. Three configurations of the prompt generator are tested: template-based descriptions, free-form descriptions incorporating diverse syntactic structures and synonym substitutions, and complex descriptions that integrate combinations of temporal actions that highlight temporal relationships. The results of the different designs are shown in Figure \ref{fig:prompts}. It can be observed that free-form descriptions, characterized by diverse syntactic structures and synonyms, significantly outperform template-based descriptions, highlighting the critical role of high-quality descriptions. Additionally, increasing the temporal complexity of the data further enhances performance.

\begin{figure}[tp]
\centering
\includegraphics[width=\linewidth]{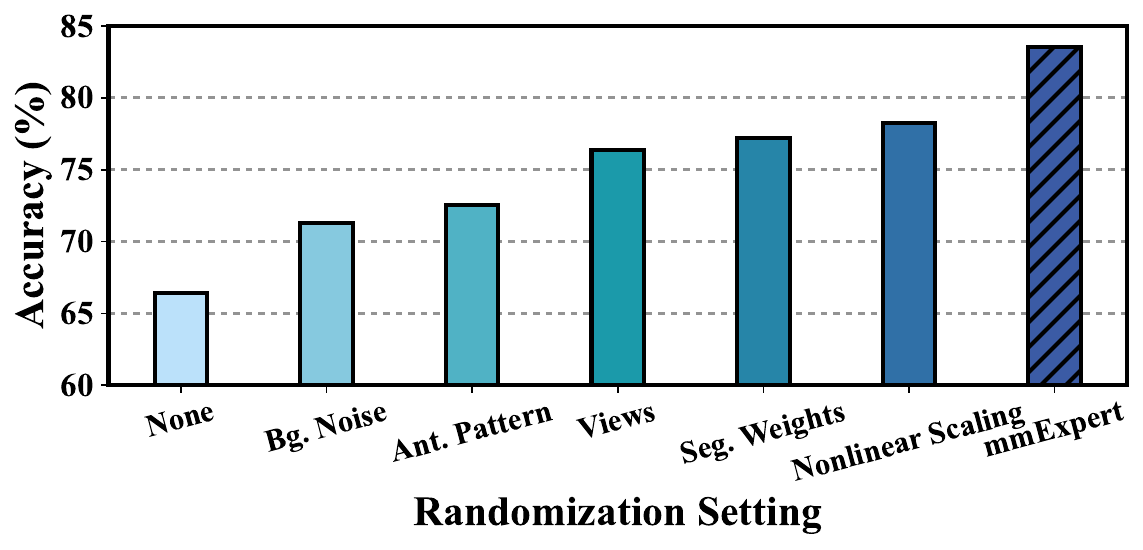}
\captionsetup{skip=0em}
\caption{Ablation study of domain randomization factors. We evaluate the final performance by assessing the impact of adding each factor individually. Legend: \systemname -- all randomization factors adopted; Bg. -- background; Ant. -- antenna; Seg. -- segment; None -- no randomization factors adopted.}
\label{fig:ablation_randomization}
\end{figure}

\noindent\textbf{Impact of domain randomization. } We conduct ablation studies to evaluate the effects of various domain randomization techniques by individually adding each randomization component. As illustrated in Figure \ref{fig:ablation_randomization}, the model achieves the best performance when all randomization factors are used. Notably, stochastic nonlinearity scaling yields the most substantial improvement. Other randomization elements, such as radar view variations, body segmentation weights, and background noise, also lead to a certain increase in performance. 

\noindent\textbf{Comparison for LLM and non-LLM models.} To evaluate our LLM's classification capability, we adapted it by replacing its final layer with a classification head and compared it against the classification model (ViT encoder + MLP decoder) previously trained on the same data for classification tasks. The results in Table \ref{tab:llm_vs_cnn} show the fine-tuned LLM classifier achieves higher accuracy. This demonstrates a key advantage: leveraging knowledge transferred from a large pre-trained model enables superior generalization and accuracy, even when fine-tuned on limited task-specific data.

\begin{table}[bp]
\centering
\captionsetup{skip=0em}
\caption{Performance comparison between our LLM-based classifier and a standard baseline.}
\label{tab:llm_vs_cnn}
\begin{tabular*}{\columnwidth}{@{\extracolsep{\fill}}lc}
\toprule
\textbf{Setting} & \textbf{Accuracy (\%)} \\
\midrule
Baseline Classifier (from scratch) & 84.6 \\
\textbf{LLM-based Classifier (fine-tuned)} & \textbf{85.1} \\
\bottomrule
\end{tabular*}
\end{table}

\subsection{Robustness to Multipath Reflections} 

To address concerns about environmental factors, we evaluated \systemname's performance in an environment with significant multipath reflections caused by nearby metallic surfaces. Our system mitigates these effects using a two-fold strategy. First, range gating isolates signals within the target's specific distance, effectively rejecting delayed multipath signals. Concurrently, by analyzing the time-Doppler signal, we filter out off-axis reflections from static objects by focusing on the target's distinct velocity profile.

This approach proved effective in our evaluation. The system achieved a classification accuracy of 84.6\% in an ideal open-space setting, which only slightly decreased to 83.4\% when strong reflectors (e.g., walls) were introduced. This minor degradation confirms the robustness of \systemname in more complex, realistic environments.

\subsection{System Overhead}

During the signal synthesis stage, with 48 GB CUDA memory (1 $\times$ Nvidia L20 GPU), the simulation of a 12-second motion (50 fps) sequence and mmWave signal synthesis can be achieved within 1 second. In the LLM inference stage, with 96 GB of CUDA memory (2 $\times$ Nvidia L20 GPUs), a single Question \& Answer (Q\&A) can be completed within 3 seconds.

\section{Related Work}
\label{sec:related_work}

\subsection{mmWave Signal Synthesis Systems} 

In recent years, numerous studies have focused on mmWave-based human sensing \cite{zhang2023survey}. However, the lack of large-scale datasets and the high cost of data collection remain significant obstacles to the development of mmWave-based human sensing systems \cite{deng2024g, xue2023towards, chen2023rf}. To overcome these challenges, researchers have proposed to synthesize mmWave radar data from diverse sources, including vision sensors~\cite{ahuja2021vid2doppler, zhang2022synthesized, xue2023towards} and text descriptions~\cite{chen2023rf, chi2024rf, zhou2024text2doppler}.

Generating mmWave radar data from text offers greater flexibility than vision-based approaches, as it is less dependent on data sources. One method uses generative models like GANs or diffusion models \cite{chi2024rf} to directly synthesize radar data, though it requires substantial real-world data for training to ensure authenticity. Another approach generates human motion sequences from text and synthesizes mmWave data via simulation \cite{chen2023rf, zhou2024text2doppler}, decoupling the process from data constraints while preserving the physical principles of mmWave radar. RFGenesis \cite{chen2023rf} pioneers this approach, combining text descriptions with diffusion models to enhance data quality. Similarly, Text2Doppler \cite{zhou2024text2doppler} employs a large corpus to guide human motion generation and subsequent radar synthesis.

\subsection{Multi-modal Large Language Models} 2D large multimodal models~\cite{liu2023visual,wang2024cogvlm} have demonstrated exceptional versatility and capability in tasks such as image captioning, visual question answering, visual grounding, visual dialogue, and even segmentation. Unified under the next-token prediction framework, these models derive significant benefits from large-scale image captioning datasets~\cite{mcoco}. Additionally, they often leverage GPT-based methods for data augmentation, achieving remarkable success within the 2D domain. This paradigm has proven both efficient and adaptable when extended to other fields, such as 3D point clouds~\cite{xu2024pointllm} and LiDAR signal interpretation~\cite{yang2025lidar}.

To the best of our knowledge, we are the first to extend large multimodal models to the domain of mmWave signal understanding by introducing \textbf{WaveLLM}, a multimodal large language model capable of directly interpreting mmWave signals. The primary challenge in this endeavor lies in the scarcity of paired mmWave-language data. To address this, we propose \systemname, a versatile data curation pipeline designed to enhance the language alignment of the mmWave encoder and facilitate the fine-tuning of WaveLLM.

\section{Conclusion}
\label{sec:conclusion}
This paper introduces \systemname, a novel framework for text-to-mmWave generation and mmWave-to-text understanding, leveraging LLMs to bridge the gap between textual descriptions and mmWave sensing data. \systemname demonstrates significant advances in scalability, cost-effectiveness, and performance. Leveraging only synthetic data, \systemname achieves state-of-the-art results in signal generation and classification tasks, showcasing robust zero-shot capabilities on real-world data. This work establishes a foundation for the integration of LLMs in mmWave sensing, paving the way for more versatile and efficient applications.

\begin{acks}
This work is supported by the National Science Fund of China under grant  No. 62472379, No. 62394341, No. 62394344, and No. 92467301, Key Research and Development Program of Zhejiang Province No. 2025C01012.
\end{acks}

\bibliographystyle{ACM-Reference-Format}
\bibliography{reference}

\end{document}